\pdfoutput=1
\documentclass[journal]{IEEEtran}
\ifCLASSINFOpdf

\else

\fi

\usepackage{amssymb}
\usepackage{amsthm}
\usepackage[cmex10]{amsmath}
\usepackage{algorithm}
\usepackage{algpseudocode}
\usepackage{graphicx}
\usepackage{xfrac}
\usepackage{subfigure}
\usepackage[export]{adjustbox}
\usepackage{makebox}
\usepackage{tikz}
\usepackage{lipsum}

\newcommand\ChapterPrecis[2]{%
	\begin{tikzpicture}[remember picture,overlay]
	\node[anchor=north,draw,yshift=-#1] at (current page.north) {\parbox[t][0.9cm][c]{2.3\linewidth}{#2}};
	\end{tikzpicture}%
}

\begin{document}
%
\title{Deep Learning of Part-based Representation of Data Using Sparse Autoencoders with Nonnegativity Constraints}


\author{Ehsan~Hosseini-Asl,~\IEEEmembership{Member,~IEEE,}
	Jacek~M.~Zurada,~\IEEEmembership{Life~Fellow,~IEEE,}
	Olfa~Nasraoui,~\IEEEmembership{Senior Member,~IEEE}
	\thanks{E.~Hosseini-Asl is with the Department
		of Electrical and Computer Engineering, University of Louisville, Louisville,
		KY, 40292 USA. (Corresponding author, e-mail: ehsan.hosseiniasl@louisville.edu).}
	\thanks{J.~M.~Zurada is with the Department
		of Electrical and Computer Engineering, University of Louisville, Louisville,
		KY, 40292 USA, and also with the Information Technology Institute, University of Social Science,\L \'{o}dz 90-113, Poland (e-mail: jacek.zurada@louisville.edu).}
	\thanks{O.~Nasraoui is with the Department 
		of Computer Science, University of Louisville, Louisville,
		KY, 40292 USA. (e-mail: olfa.nasraoui@louisvile.edu).}%
}

\maketitle

\begin{abstract}
	We demonstrate a new deep learning autoencoder network, trained by a nonnegativity constraint algorithm (NCAE), that learns features which show part-based representation of data. The learning algorithm is based on constraining negative weights. The performance of the algorithm is assessed based on decomposing data into parts and its prediction performance is tested on three standard image data sets and one text dataset. The results indicate that the nonnegativity constraint forces the autoencoder to learn features that amount to a part-based representation of data, while improving sparsity and reconstruction quality in comparison with the traditional sparse autoencoder and Nonnegative Matrix Factorization. It is also shown that this newly acquired representation improves the prediction performance of a deep neural network. 
\end{abstract}


\IEEEpeerreviewmaketitle

\ChapterPrecis{1cm}{\textcopyright 2015 IEEE. Personal use of this material is permitted. Permission from IEEE must be obtained for all other uses, in any current or future media, including reprinting/republishing this material for advertising or promotional purposes, creating new collective works, for resale or redistribution to servers or lists, or reuse of any copyrighted component of this work in other works.}

\section{Introduction}
\IEEEPARstart{R}{ecent} studies have shown that deep architectures are capable of learning complex data 
distributions while achieving good generalization performance and efficient representation of patterns in challenging recognition tasks \cite{bengio2007scaling,bengio2009learning,hinton2006reducing,deng2014tutorial,bengio2013guest,hutchinson2013tensor}. Deep architecture networks have many levels of nonlinearities, giving them an ability
to compactly represent highly nonlinear complex mappings. However, they are difficult to train, since there are many hidden layers with many connections, which causes gradient-based optimization with random initialization to get stuck in poor solutions \cite{bengio2007greedy}. To improve on this bottleneck, a greedy layer-wise training algorithm 
was proposed in \cite{hinton2006fast}, where each layer is separately initialized by unsupervised pre-training, then the stacked layers are fine-tuned using a supervised learning algorithm  \cite{hinton2006reducing,bengio2007greedy}. It was shown that an unsupervised pre-training phase of each layer helps in capturing the patterns 
in high-dimensional data, which results in a better representation in a low-dimensional encoding space \cite{hinton2006reducing}, and could result in more sparse feature learning \cite{marc2007sparse}. This pre-training also improves the supervised fine-tuning algorithm for classification by guiding the learning algorithm towards local minima of the error function, that support better generalization on training data \cite{raina2007self,erhan2010does}.

There are two popular algorithms for unsupervised learning which have been shown to work well for producing
a good representation for initializing deep structures \cite{vincent2008extracting}: Restricted Boltzmann Machines (RBMs), trained 
with contrastive divergence \cite{hinton2002training}, and different types of autoencoders \cite{bengio2009learning}. In this paper, we focus on unsupervised feature learning based on autoencoders.

Autoencoder neural networks are trained with an unsupervised learning algorithm based on reconstructing the input from its encoded representation, while constraining the representation to have some desirable properties. Deep networks based on autoencoders are created by stacking pre-trained autoencoders layer by layer, followed by a supervised fine-tuning algorithm \cite{bengio2007greedy}. As discussed before, a key contributor to successful
training of a deep network is a proper initialization of the layers based on a local unsupervised 
criterion \cite{vincent2008extracting}. In the case of autoencoders, the reconstruction error is used as a local criterion for unsupervised learning of each layer. Each layer produces a representation of its input at the hidden layer, where this representation is used as the input to the next layer. Therefore,
a learning algorithm which results in lower reconstruction error at each layer should create a more accurate representation of data, and deliver a better initialization of layer parameters; This, in turn, improves the deep network's prediction performance \cite{vincent2008extracting}. One additional criterion proposed for this model is sparsity of the autoencoding. Sparseness of the representation, i.e. activity of hidden nodes, has been shown to have some advantages in
robustness to noise and improved classification in high-dimensional spaces  \cite{marc2007sparse,lee2007sparse}.  

This paper demonstrates how to achieve a meaningful representation from data that discovers the hidden structure of high-dimensional data based on autoencoders \cite{bengio2013representation}. It has been shown that data is represented in hierarchical layers through the visual cortex~\cite{van1994neural}, where some psychological and physiological evidence showed that data is represented by part-based decomposition in the human brain~\cite{wachsmuth1994recognition}. Inspired by the idea of sparse coding \cite{olshausen1997sparse,poultney2006efficient,makhzani2013k} and Nonnegative Matrix Factorization (NMF) \cite{olshausen1996emergence,lee1999learning}, learning features that exhibit sparse part-based representation of data (decomposing data into parts) is expected to disentangle the hidden structure of data. We develop a deep network by extracting part-based features in hierarchical layers, and show that these features result in good generalization ability for the trained model, and improve the reconstruction error. Using NMF, the features and the encoding of data are forced to be nonnegative, which results in part-based additive representation of data. However, while sparse coding within NMF needs an
expensive optimization process to find the encoding of test data, this process is very fast in autoencoders \cite{marc2007sparse}. Therefore, training an autoencoder which could exploit the benefits of part-based representation using nonnegativity is expected to improve the performance of a deep learning network. 


The most closely related work to ours is that of Lemme et al. on the Nonnegative Sparse Autoencoder (NNSAE)\cite{lemme2012online}. In their approach, an online training algorithm has been developed for an autoencoder with tied weights, and a linear function at the output layer of the autoencoder to simplify the training algorithm. The key difference compared to our network is that we use a general autoencoder with trainable weights for both the hidden and the output layers. We also use a nonlinear function for the nodes in each layer which makes our model more flexible. The other related work is by Chorowski et al. \cite{chorowski2014learning}, who demonstrated that a multilayer perceptron network with softmax output nodes trained with nonnegative weights, was capable of extracting understandable latent features, which consist of part-based representation of data. It was illustrated with extracting characteristic parts of handwritten digits and extracting semantic features from data. However, the improved transparency had slightly reduced the network's prediction performance. Moreover, it is difficult to apply the method to train a large deep network with several layers. The reason is that random initialization of the network was used instead of pretraining with the greedy layer-wise algorithm \cite{chorowski2014learning}. In contrast, our method for pretraining and fine-tuning of a deep network can be easily scaled up to large deep networks, since it uses greedy layer-wise pretraining. Another related work is by Nguyen et al. \cite{nguyen2013learning}, with the nonnegativity constraint applied to train an RBM network (NRBM). It possesses a certain similarity to RBM and NMF in terms of part-based representation and classification accuracy. However, RBM uses a stochastic approach based on maximizing the joint probability between the visible and hidden units, whereas the autoencoder uses a deterministic approach based on minimizing the reconstruction error. In this case, part-whole decomposition of data based on additive parts can be easily translated by incorporating nonnegative weights in the encoding and decoding layers of the autoencoder, which tries to minimize the reconstruction error. Therefore, we use an autoencoder instead of RBM to produce a part-based representation of data. 

\begin{figure*}[htb!]
	
	\begin{minipage}[b]{0.5\linewidth}
		\centering
		\includegraphics[scale=0.4]{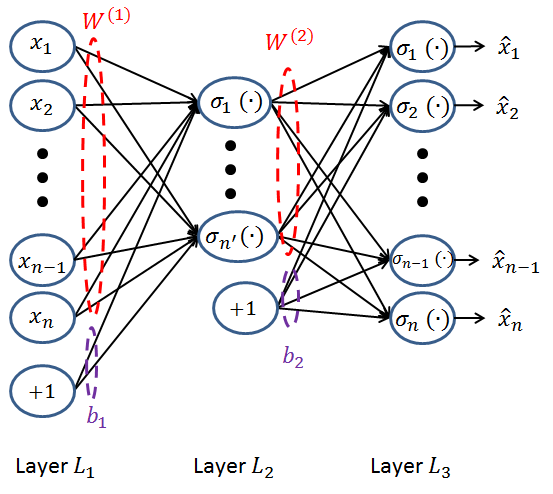}
		{{\footnotesize (a)}}
	\end{minipage}
	\begin{minipage}[b]{0.5\linewidth}
		\centering
		\includegraphics[scale=0.4]{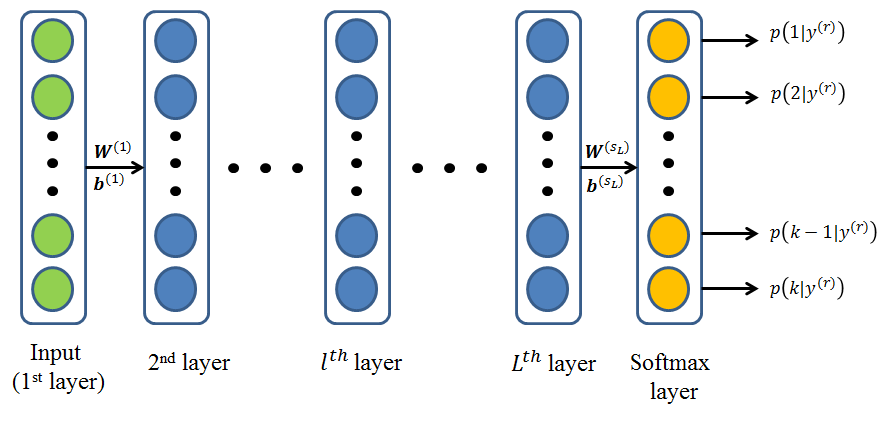}
		{{\footnotesize(b)}}
	\end{minipage}
	\caption{(a) Schematic diagram of (a) a three-layer autoencoder and (b) a deep network}
		\label{fig:diagram}
\end{figure*}

In this paper, we propose a new approach to train an autoencoder by introducing a nonnegativity constraint into its learning to learn a sparse, part-based representation of data. The training is then extended to train a deep network with stacked autoencoders and a softmax classification layer, while again constraining the weights of the network to be nonnegative. Our goal is two-fold: part-based representation in the autoencoder network to improve its ability to disentangle the hidden structure of the data, and producing a better reconstruction of the data. We also show that these criteria improve the prediction performance of a deep learning network.

The performance of our method is first compared with Sparse Autoencoder (SAE) \cite{ng2011ufldl}, Nonnegative Sparse Autoencoder (NNSAE) \cite{lemme2012online}, and NMF \cite{lee2000algorithms} in terms of extracting latent features, reconstruction error and sparsity of representation on several image and text benchmark datasets. Then the classification performance of our deep learning approach is shown to significantly outperform the related deep networks reported in the literature based on SAE, NNSAE, Denoising Autoencoder (DAE) \cite{vincent2008extracting}, and Dropout Autoencoder (DpAE) \cite{hinton2012improving}.


\section{Methods}\label{sec:methods}
As mentioned, an autoencoder neural network tries
to reconstruct its input vector at the output through unsupervised learning~\cite{vincent2008extracting,hinton1994autoencoders}. As shown in Fig. \ref{fig:diagram}(a), it tries to learn a function, 
\begin{equation}
\hat{\textbf{x}}=f_{\textbf{W},\textbf{b}}(\textbf{x})\approx \textbf{x} 
\label{eq:autoencoder}
\end{equation}
where $\textbf{x}$ is the input vector, while $\textbf{W}=\left\{\textbf{W}_{1},\textbf{W}_{2}\right\}$ 
and $\textbf{b}=\left\{\textbf{b}_{1},\textbf{b}_{2}\right\}$ represent weights and biases of both layers, respectively. 
It takes an input vector $\textbf{x}\in [0,1]^{n}$, and first maps it to a hidden
representation through a deterministic mapping, parametrized by $\bf{\theta}_{1}=\left\{\textbf{W}_{1},\textbf{b}_{1}\right\}$, and given by
\begin{equation}
\textbf{h}=g_{\theta_{1}}(\textbf{x})=\sigma\left(\textbf{W}_{1} \textbf{x}+\textbf{b}_{1}\right)
\end{equation}
where $\textbf{h}\in [0,1]^{n'}$, $\textbf{W}_{1} \in R^{n'\times n}$, $\textbf{b}\in R^{n'\times 1}$, and $\sigma(x)$ denotes an element-wise application of the logistic sigmoid, $\sigma(x)=\sfrac{1}{(1+exp(-x))}$. The resulting hidden representation,
$\bf{h}$, is then mapped back to a reconstructed vector, $\hat{\textbf{x}}\in [0,1]^{n}$, by a similar mapping function, parametrized by $\theta_{2}=\left\{\textbf{W}_{2},\textbf{b}_{2}\right\}$,
\begin{equation}
\hat{\textbf{x}}=g_{\theta_{2}}(\textbf{h})=\sigma\left(\textbf{W}_{2} \textbf{h}+\textbf{b}_{2}\right)
\end{equation}
where $\textbf{W}_{2} \in R^{n\times n'}$ and $\textbf{b}_{2}\in R^{n\times 1}$. To optimize the 
parameters of the model in Eq. (\ref{eq:autoencoder}), i.e. $\theta=\left\{\theta_{1},\theta_{2}\right\}$, 
the average reconstruction error is used as the cost function,

\begin{equation}
\label{eq:cost}
J_{\text{E}}(\textbf{W},\textbf{b})=\frac{1}{m}\sum_{r=1}^{m}\frac{1}{2}\parallel \hat{\textbf{x}}^{(r)}-\textbf{x}^{(r)}\parallel^{2}
\end{equation}
where $m$ is the number of training samples.

By imposing meaningful limitations on parameters $\theta$, e.g. limiting the dimension $n'$ of the hidden representation $\textbf{h}$, the autoencoder learns a
compressed representation of the input, which helps discover the latent structure of data in a high-dimensional space. 

Sparse representation can provide a simple interpretation of the
input data in terms of a reduced number of parts and by extracting the structure hidden in the data. 
Several algorithms were proposed to learn a sparse representation using autoencoders \cite{hinton2006fast,poultney2006efficient}. One common method for imposing sparsity is to limit the activation of hidden units $\textbf{h}$ using the Kullback-Leibler (KL) divergence function \cite{lee2007sparse,nair20093d}. Let $h_{j}\left(\textbf{x}^{(r)}\right)$ denote the activation of hidden unit $j$ with respect to the input $\textbf{x}^{(r)}$. Then the average activation of this hidden unit is:
\begin{equation}
\hat{p}_{j}=\frac{1}{m}\sum_{r=1}^{m}h_{j}(\textbf{x}^{(r)})
\end{equation}
To enforce sparsity, we constrain the average activation $\hat{p}_{j}=p$, where $p$ is the sparsity
parameter chosen to be a small positive number near $0$. This also relates to the normalization of the input to the neurons of the next layer which results in faster convergence of training using the backpropagation algorithm \cite{lecun2012efficient}. To use this constraint in Eq. (\ref{eq:cost}), 
we try to minimize the KL divergence similarity between $\hat{p}_{j}$ and $p$,
\begin{equation}
\label{eq:cost_kl}
J_{\text{KL}}(p\parallel \hat{\textbf{p}})=\sum_{j=1}^{n'}p \log\frac{p}{\hat{p}_{j}}+(1-p)\log\frac{1-p}{1-\hat{p}_{j}}
\end{equation}
where $\hat{\textbf{p}}$ is the vector of average hidden activities. To prevent overfitting, a weight decay term is also added to the 
cost function of Eq. (\ref{eq:cost}) \cite{moody1995simple}. The final cost function for learning a Sparse Autoencoder (SAE) becomes as follows:

\begin{equation}
\begin{aligned}
J_{\text{SAE}}(\textbf{W},\textbf{b})&=J_{\text{E}}(\textbf{W},\textbf{b})+\beta J_{\text{KL}}(p\parallel \hat{\textbf{p}})\\
&+\frac{\lambda}{2}\sum_{l=1}^{2}\sum_{i=1}^{s_{l}}\sum_{j=1}^{s_{l+1}}\left(w_{ij}^{(l)}\right)^{2}
\end{aligned}
\label{eq:cost:sparse}
\end{equation}
where $\beta$ controls the sparsity penalty term, $\lambda$ controls the penalty term facilitating weight decay, and $s_{l}$ and $s_{l+1}$ are the sizes of adjacent layers. 

\subsection{Part-based Representation Using a Nonnegativity Constrained Autoencoder (NCAE)}
Ideally, part-based representation is implemented through decomposing data into parts, which when combined,
produce the original data. However, the combination of parts here is only allowed to be additive \cite{lee1999learning}. As shown in \cite{chorowski2014learning} that demonstrates part-based representation, the input data can be decomposed in each layer into parts, while the weights in $\textbf{W}$ are constrained to be nonnegative \cite{chorowski2014learning}.

Intuitively, to improve the performance of the autoencoder in terms of reconstruction of input data, it should be able to decompose data into parts which are sparse in the encoding layer and then combine them in an additive manner in the decoding layer. To achieve this goal, a nonnegativity constraint is imposed on
the connecting weights $\textbf{W}$. This means that the column vectors of $\textbf{W}$ are coerced to be sparse, i.e. only a fraction of entries per column should remain non-zero. 

To encourage nonnegativity in $\textbf{W}$, the weight decay term in Eq. (\ref{eq:cost:sparse}) is replaced and a quadratic function \cite{wright1999numerical,nguyen2013learning} is used. This results in the following cost function for NCAE:

\begin{equation}
\begin{aligned}
J_{\text{NCAE}}\left(\textbf{W},\textbf{b}\right)&=J_{\text{E}}\left(\textbf{W},\textbf{b}\right)+\beta J_{\text{KL}}(p\parallel \hat{\textbf{p}})\\
&+\frac{\alpha}{2}\sum_{l=1}^{2}\sum_{i=1}^{s_{l}}\sum_{j=1}^{s_{l+1}}f\left(w_{ij}^{(l)}\right)
\end{aligned}
\label{eq:cost:nonneg}
\end{equation}
where 
\begin{equation}
f(w_{ij}) = \left\{ 
\begin{array}{l l}
w_{ij}^{2} & \quad w_{ij}<0\\
0 & \quad w_{ij}\geq0
\end{array} \right.
\label{eq:constraint}
\end{equation}
and $\alpha\geq0$. Minimization of Eq. (\ref{eq:cost:nonneg}) would result in reducing the 
average reconstruction error, increased sparsity of hidden layer activations, and reduced number of nonnegative weights of each layer.

To update the weights and biases, we compute the gradient of Eq. (\ref{eq:cost:nonneg}) used in the backpropagation algorithm:

\begin{equation}
w_{ij}^{(l)} =w_{ij}^{(l)}-\eta \frac{\partial}{\partial w_{ij}^{(l)}}J_{\text{NCAE}}(\textbf{W},\textbf{b})
\label{eq:w-update}
\end{equation}
\begin{equation}
b_{i}^{(l)}=b_{i}^{(l)}-\eta \frac{\partial}{\partial b_{i}^{(l)}}J_{\text{NCAE}}(\textbf{W},\textbf{b})
\label{eq:b-update}
\end{equation}
where $\eta>0$ is the learning rate. The derivative of Eq. (\ref{eq:cost:nonneg}) with respect to the weights consists of three terms as shown below,

\begin{equation}
\begin{aligned}
\frac{\partial}{\partial w_{ij}^{(l)}}J_{\text{NCAE}}(\textbf{W},\textbf{b})&=\frac{\partial}{\partial w_{ij}^{(l)}}J_{\text{E}}\left(\textbf{W},\textbf{b}\right)\\
&+\beta \frac{\partial}{\partial w_{ij}^{(l)}}J_{\text{KL}}\left(p\parallel \hat{\textbf{p}}\right)\\
&+\alpha g\left(w_{ij}^{(l)}\right)
\end{aligned}
\label{eq:grad-NCAE}
\end{equation}
where
\begin{equation}
g(x) = \left\{ 
\begin{array}{l l}
w_{ij} & \quad w_{ij}<0\\
0 & \quad w_{ij}\geq0
\end{array} \right.
\label{eq:grad-constraint}
\end{equation}
The derivative term in Eq. (\ref{eq:b-update}) and the first two terms in Eq. (\ref{eq:grad-NCAE}) are computed using the backpropagation 
algorithm \cite{ng2011ufldl, zurada1992introduction}. 

\begin{figure*}[htb!]
	
	\begin{minipage}[b]{1.0\linewidth}
		\centering
		\includegraphics[width=18cm]{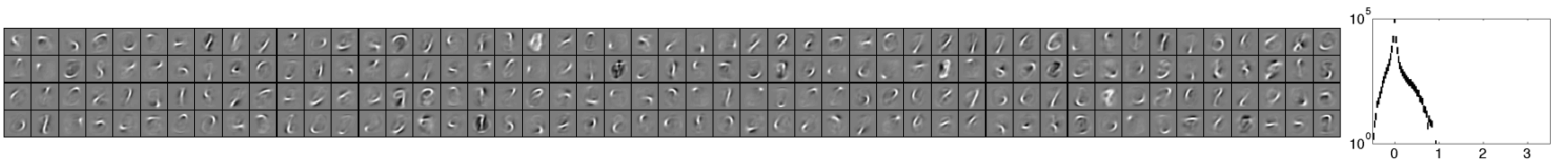}
		\vspace{-2mm}
		{{\footnotesize (a) SAE}}
	\end{minipage}
	
	\begin{minipage}[b]{1.0\linewidth}
		\centering
		\vspace{1mm}
		\includegraphics[width=18cm]{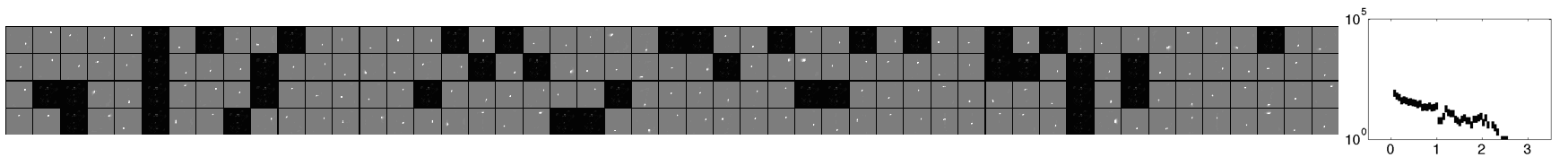}
		\vspace{-2mm}
		{{\footnotesize(b) NNSAE}}
	\end{minipage}
	
	\begin{minipage}[b]{1.0\linewidth}
		\centering
		\vspace{1mm}
		\includegraphics[width=18cm]{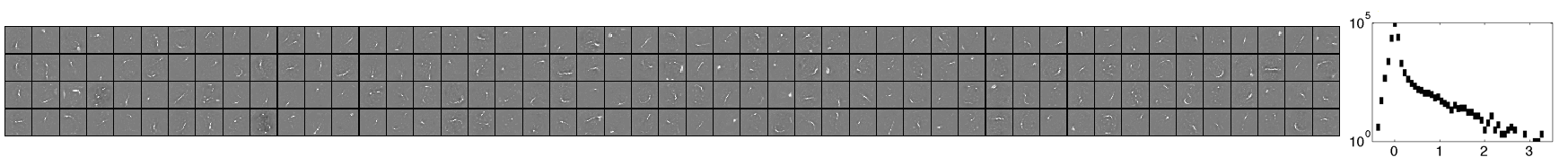}
		\vspace{-2mm}
		{{\footnotesize(c) NCAE*}}
	\end{minipage}
	\begin{minipage}[b]{1.0\linewidth}
		\centering
		\vspace{1mm}
		\includegraphics[width=18cm]{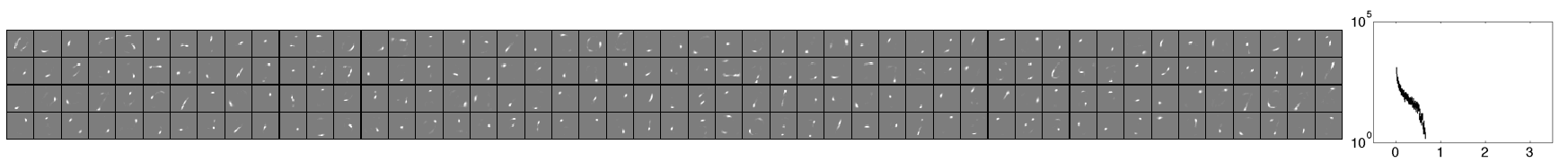}
		\vspace{-2mm}
		{{\footnotesize(d) NMF}}
	\end{minipage}
	\caption{196 Receptive fields ($\mathbf{W}^{(1)}$) with weight histogram learned from MNIST digit data set using (a) SAE, (b) NNSAE, (c) NCAE*, and (d) NMF. Black pixels indicate negative, and white pixels indicate positive weights. Black nodes in (b) indicate neurons with zero weights. The range of weights are scaled to [-1,1] and mapped to the the graycolor map. $w<=-1$ is assigned to black, and $w>=1$ is assigned to white color.}
	\label{fig:mnist-receptivefield}
\end{figure*}

\begin{figure*}[htb!]
	
	\begin{minipage}[b]{1.0\linewidth}
		\centering
		\includegraphics[width=18cm]{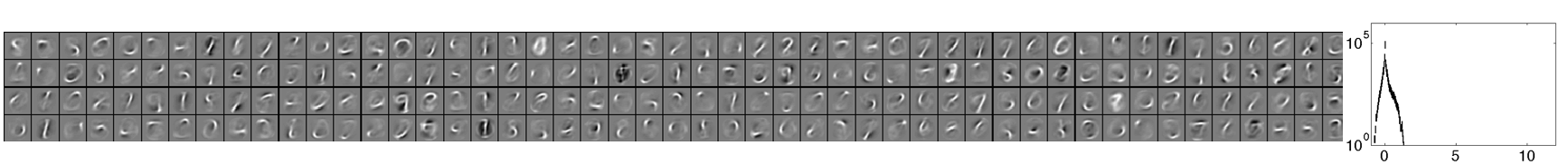}
		\vspace{-2mm}
		{{\footnotesize (a) SAE}}
	\end{minipage}
	
	\begin{minipage}[b]{1.0\linewidth}
		\centering
		\vspace{1mm}
		\includegraphics[width=18cm]{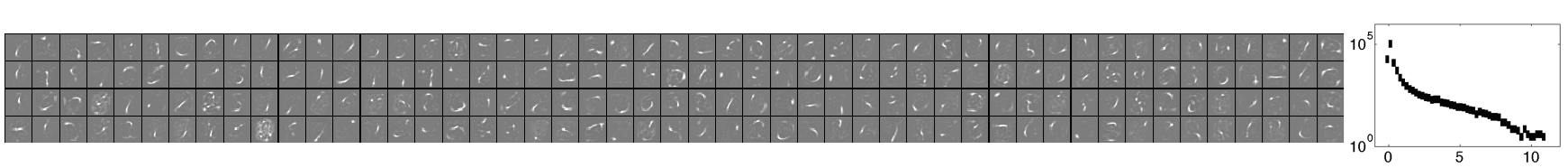}
		\vspace{-2mm}
		{{\footnotesize(b) NCAE*}}
	\end{minipage}
		\caption{196 decoding filters ($\mathbf{W}^{(2)}$) with weight histogram learned from MNIST digit data set using (a) SAE and (b)  NCAE*. Black pixels indicate negative, and white pixels indicate positive weights. Black nodes in (b) indicate neurons with zero weights.}
		\label{fig:mnist-decodingfilter}
\end{figure*}
	
\subsection{Deep Learning using the Nonnegative Constrained Autoencoder (NCAE)}
\vspace{-1mm}
A greedy layer-wise training algorithm is used to build a deep network, with each layer pre-trained separately by unsupervised feature learning \cite{hinton2006fast}. In this paper, a deep NCAE network is pretrained, i.e., several layers of the autoencoder are trained step by step, with the hidden activities of the previous stage used as input for the next stage. Finally, the hidden activities of the last autoencoder is used as an input to a softmax regression classifier to be trained in a supervised mode. In our approach, the nonnegative weights of the softmax classifier during training are constrained as described for training NCAE. The misclassification cost function of the softmax classifier is

\small
\begin{equation}
J_{\textit{CL}}\left(\textbf{W}\right)=-\frac{1}{m}\left[\sum_{r=1}^{m}\sum_{p=1}^{k}1\left(y^{(r)}=p\right)log\frac{e^{\textbf{w}_{p}^{T}x^{(r)}}}{\sum_{l=1}^{k}e^{\textbf{w}_{l}^{T}x^{(r)}}}\right]
\label{eq:cost-misClassification}
\end{equation}
\normalsize
where $m$ is the number of samples, $k$ is the number of classes, $\bf{W}$ is the matrix of input weights of all nodes in the softmax layer, and $\textbf{w}_{p}$ is the $p$-th column of $\bf{W}$ referring to the input weights of the $p$-th softmax node. Therefore, we define the cost function of Nonnegativity-Constrained Softmax as
\begin{equation}
J_{\textit{NC-Softmax}}\left(\textbf{W}\right)=J_{\textit{CL}}\left(\textbf{W}\right)
+\frac{\alpha}{2}\sum_{i=1}^{s_{L}}\sum_{j=1}^{k}f\left(w^{(L)}_{ij}\right)
\label{eq:cost-NCsoftmax}
\end{equation}
where $s_{L}$ denotes the number of hidden nodes of the final autoencoder, $f(\cdot)$ is as in Eq. (\ref{eq:constraint}) to penalize the negative weights of the softmax layer.
The final step of greedy-wise training is to stack the trained NCAE and softmax layers, and fine-tune the network in supervised mode for best classification accuracy~\cite{hinton2006fast}. Only the negative weights of the softmax layer are constrained during fine-tuning. The cost function for fine-tuning the Deep Network (DN) is given by

\begin{equation}
J_{\textit{DN}}\left(\textbf{W},\textbf{b}\right)=J_{\textit{CL}}\left(\textbf{W}_{DN},\textbf{b}_{DN}\right)
+\frac{\alpha}{2}\sum_{i=1}^{s_{L}}\sum_{j=1}^{k}f\left(w^{(L)}_{ij}\right)
\label{eq:cost-FineTuning}
\end{equation}
where $\textbf{W}_{DN}$ contains the input weights of the NCAE and softmax layers, and $\textbf{b}_{DN}$ is the bias input of NCAE layers, as shown in Fig. \ref{fig:diagram}(b).

A batch gradient descent algorithm is used, where the Limited-memory BFGS (L-BFGS) quasi-Newton method \cite{byrd1995limited} is employed for 
minimization of Eq. (\ref{eq:cost:nonneg}), Eq. (\ref{eq:cost-NCsoftmax}), and Eq. (\ref{eq:cost-FineTuning}). The L-BFGS algorithm computes an approximation of the inverse of the Hessian matrix, which results in less memory to store the vectors which approximate the Hessian matrix. The details of the algorithm and the software implementation can be found in \cite{schmidt2008}.

\section{Experimental Results}\label{sec:experiments}
This section reports the performance tests of the proposed method in unsupervised feature learning for three benchmark image data sets and one text dataset. A deep network using NCAE as a building block is trained, and its classification performance is evaluated. We use the MNIST digit data set for handwritten digits~\cite{lecun1998gradient}, the ORL face data set~\cite{samaria1994parameterisation} for face images, and the small NORB object recognition dataset~\cite{lecun2004learning}. The Reuters 21578 document corpus is also used to evaluate the ability of the proposed method in learning semantic features. The Matlab implementation of the NCAE algorithm can be downloaded from 


\begin{table}[htb!]
	\centering
	\scriptsize
	\caption{Parameter settings of each algorithm}
	\begin{tabular}{c c c c}
		\hline
		\text{Parameters} & \text{SAE} & \text{NCAE*} & \text{NMF} \\
		\hline
		\text{Sparsity penalty} ($\beta$) & 3 &	3 &	\text{-} \\
		\hline
		\text{Sparsity parameter (p)} & 0.05 &	0.05 &	\text{-} \\
		\hline
		\text{Weight decay penalty} ($\lambda$) & 0.003 &	\text{-} &	\text{-} \\
		\hline
		\text{Nonnegativity constraint penalty} ($\alpha$) & \text{-} &	0.003 &	\text{-} \\
		\hline
		\text{Convergence Tolerance} & 1e-9 &	1e-9 &	1e-9 \\
		\hline
		\text{Maximum No. of Iterations} & 400 &	400 &	400 \\
		\hline
		\label{table:parameters}
	\end{tabular}
\end{table}

\begin{figure}[htb!]
	\centering
	\includegraphics[width=8.5cm]{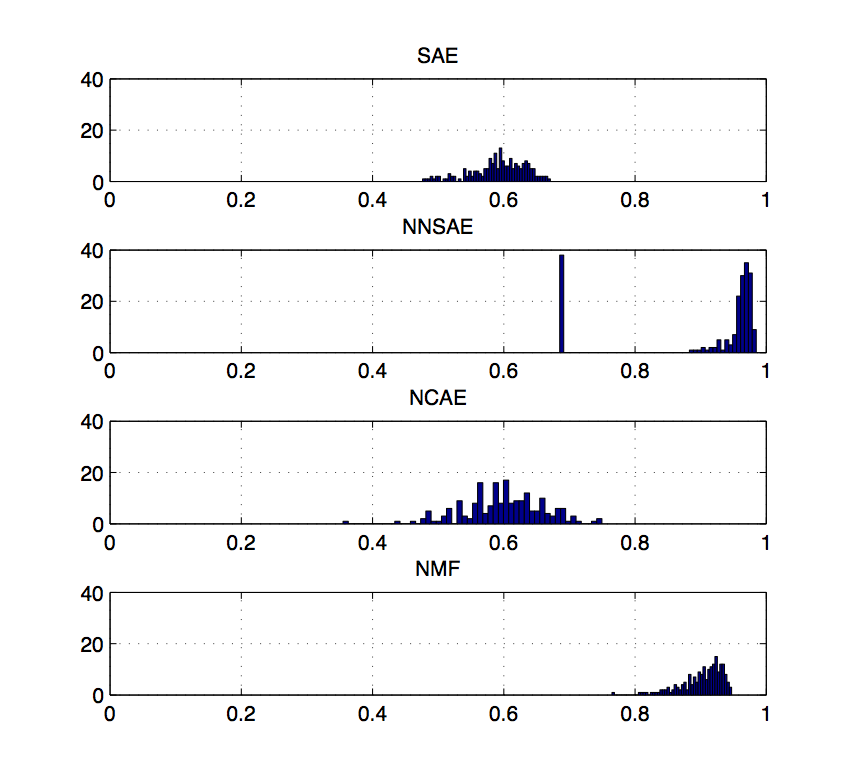}
	\caption{Histogram of the sparseness criterion~\cite{hoyer2004non} measured on 196 receptive fields.}
	\label{fig:mnist-w1-sparsity} 
\end{figure} 

\begin{figure}[htb!]
	\centering
	\includegraphics[width=8.5cm]{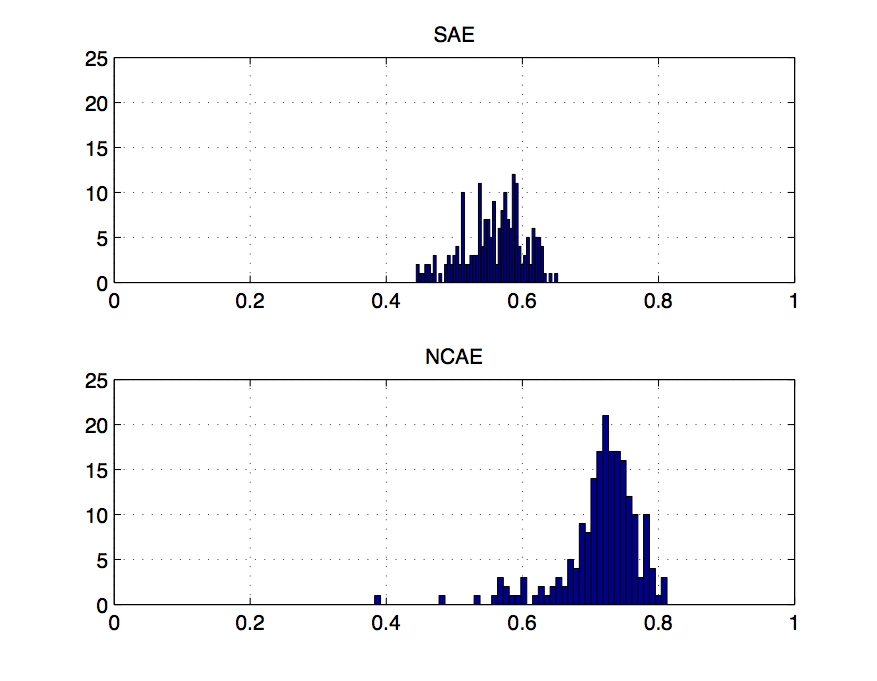}
	\caption{Histogram of the sparseness criterion~\cite{hoyer2004non} measured on 196 decoding filters.}
	\label{fig:mnist-w2-sparsity} 
\end{figure}

\begin{figure*}[t]
	\begin{minipage}[b]{0.5\linewidth}
		\centering
		\includegraphics[width=9cm]{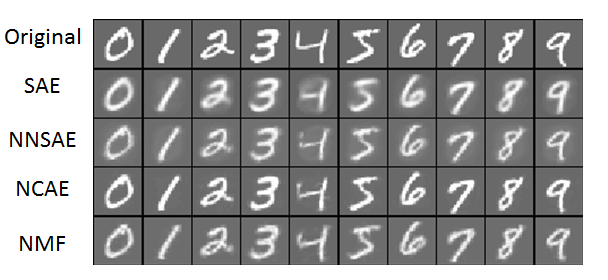}
		{{\footnotesize(a) Original and Reconstruction}}
		\label{fig:mnist-reconstruction}
	\end{minipage}
	\begin{minipage}[b]{0.5\linewidth}
		\centering
		\includegraphics[width=8cm]{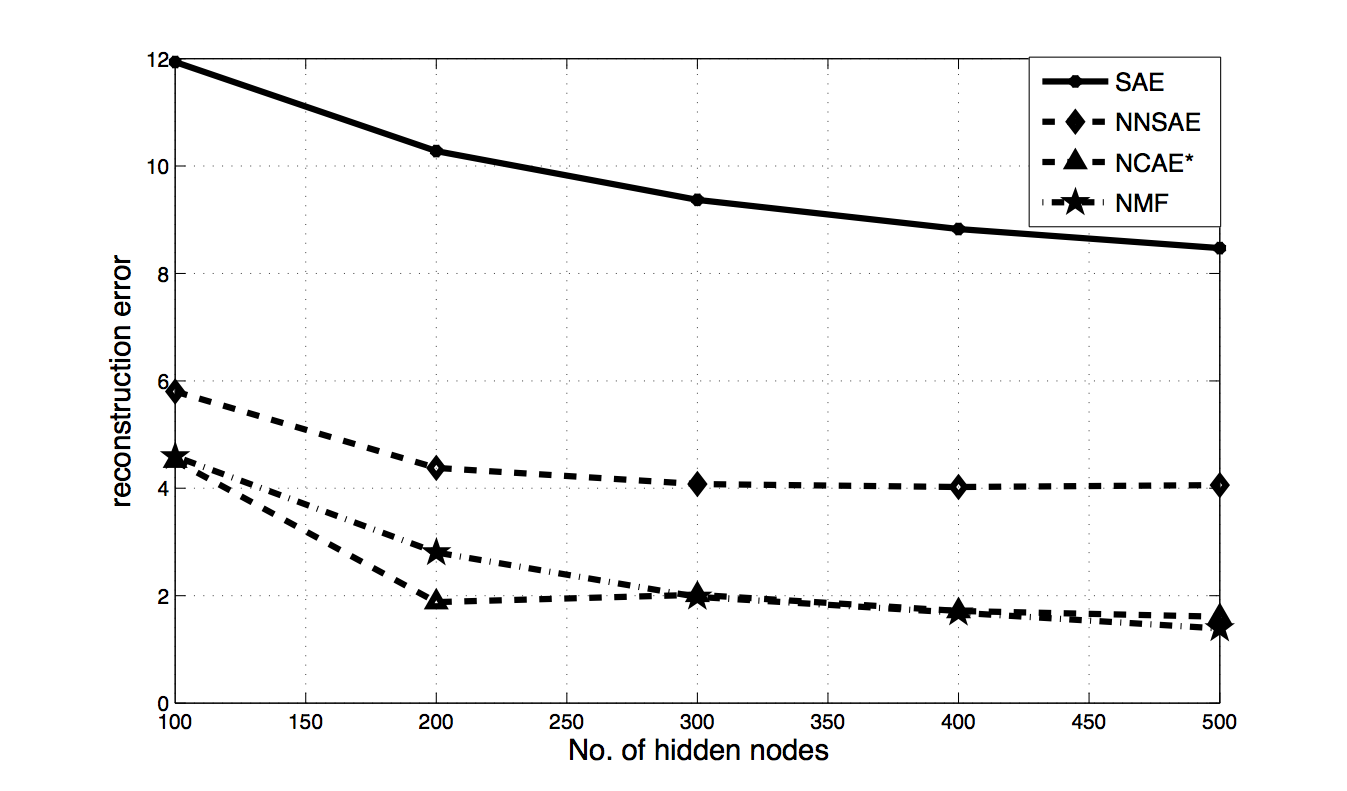}
		{{\footnotesize (b) Reconstruction Error}}
		\label{fig:mnist-reconsterror}
	\end{minipage}
	\caption{Performance comparison, (a) reconstruction of the MNIST digits data set by 196 receptive fields, using SAE (error=7.5031), NNSAE\cite{lemme2012online} (error=4.3779), NCAE* (error=1.8799), and NMF (error=2.8060), (b) reconstruction error computed by Eq. (\ref{eq:cost}). The performance is computed on test data.}
	\label{fig:MNIST-performancecomparison}
\end{figure*}

\subsection{Unsupervised Feature Learning}
A three-layer NCAE network using Eq.(\ref{eq:cost:nonneg}) was trained. In the case of image data, the input weights of hidden nodes $\textbf{W}_{1}$ are rendered as images called receptive fields. The results of the NCAE method are compared to the receptive fields learned by a three-layer Sparse Autoencoder (SAE) of Eq. (\ref{eq:cost:sparse}), Nonnegative Sparse Autoencoder (NNSAE) \cite{lemme2012online}, and the basis images learned by NMF. The multiplicative algorithm has been used to compute the basis images $\bf{W}$ of NMF~\cite{lee1999learning}. 
In the case of text data, $\textbf{W}_{1}$ represents the group of words to evaluate the ability to extract meaningful features connected to the topics in the document corpus.

To tune the hyperparameters, each algorithm has been tested with a range of values for each regularization parameter to minimize the cost in Eq. (\ref{eq:cost}) and Eq. (\ref{eq:cost:nonneg}). The value for each parameter is shown in Table \ref{table:parameters}\footnote{These were set after trial and error.}. The NNSAE training parameters are set as described in \cite{lemme2012online}.

\subsubsection{Learning Part-based Representation of Images}
In the first experiment, an NCAE network was trained on the MNIST digit data set. This dataset contains $60,000$ training and $10,000$ testing grayscale images of handwritten digits, scaled and centered inside a $28\times 28$ pixel box. The NCAE network contains $196$ nodes in the hidden layer. Its receptive fields have been compared with those of SAE, NNSAE, and NMF basis images in Fig.~\ref{fig:mnist-receptivefield}, and decoding filters are compared with SAE in Fig.~\ref{fig:mnist-decodingfilter}, with the histogram of weights for each algorithm. The results show that receptive fields, learned by NCAE, are more sparse and localized than SAE, NNSAE, and NMF. The darker pixels in SAE features indicate negative input weights. In contrast, those values 
are reduced in NCAE features due to the nonnegativity constraint. Features, learned by NCAE in Fig.~\ref{fig:mnist-receptivefield} indicate that basic structures of handwritten digits such as strokes and dots are discovered, whereas these are much less visible in SAE, where some features are parts of digits or the whole digits in a blurred form. On the other hand, the features learned by NNSAE and NMF are more local than NCAE, since it is harder to judge them as strokes and dots or parts of digits. As a result, Fig. \ref{fig:mnist-receptivefield} and Fig.~\ref{fig:mnist-decodingfilter} indicate that the NCAE network learns a sparse and part-based representation of handwritten digits that is easier to interpret, by constraining the negative weights as demonstrated by the weight histogram. To better investigate the sparsity of weights in the NCAE network, the sparseness is measured using the relationship between the $\ell_{1}$ and $\ell_{2}$ norms proposed in~\cite{hoyer2004non}, and the sparseness histograms are compared with other methods in Fig.~\ref{fig:mnist-w1-sparsity} and Fig.~\ref{fig:mnist-w2-sparsity}, for the receptive fields and decoding filters, respectively. The results indicate that the nonnegativity constraints improve the sparsity of weights in the encoding and decoding layer.

To evaluate the performance of NCAE in terms of digit reconstruction, the selected reconstructed digits and the reconstruction error of NCAE for different numbers of hidden nodes are compared with those of SAE, NNSAE, and NMF in Fig. \ref{fig:MNIST-performancecomparison}.
The reconstruction of ten selected digits from ten classes is shown in Fig. \ref{fig:MNIST-performancecomparison}(a). The top row depicts the original digits from the data set, where the reconstructed digits 
using SAE, NNSAE, NCAE, and NMF algorithms are shown below. It is clear that the digits
reconstructed by NCAE are more similar to the original digits than those by the SAE and NNSAE methods, and also contain fewer errors. On the other hand, the results of NCAE and NMF are similar, while digits in NMF are more blurred than NCAE, which indicates reconstruction errors. In order to test the performance of our method using different numbers of hidden neurons, the reconstruction error (Eq. \ref{eq:cost}) of all digits of the MNIST data set is depicted in Fig. \ref{fig:MNIST-performancecomparison}(b). The results
demonstrate that NCAE outperforms SAE and NNSAE for different numbers of hidden neurons. It can be seen that the reconstruction errors in NCAE and NMF methods are the lowest and similar, whereas NCAE shows better 
reconstruction over NMF in one case. The results in Fig. \ref{fig:MNIST-performancecomparison}(b) demonstrate that
the nonnegativity constraint forces the autoencoder networks to learn part-based representation of digits, i.e. strokes and dots, and it results in more accurate reconstruction from their encodings than SAE and NNSAE.

\begin{figure}[htb!]
	\centering
	\includegraphics[width=7.5cm]{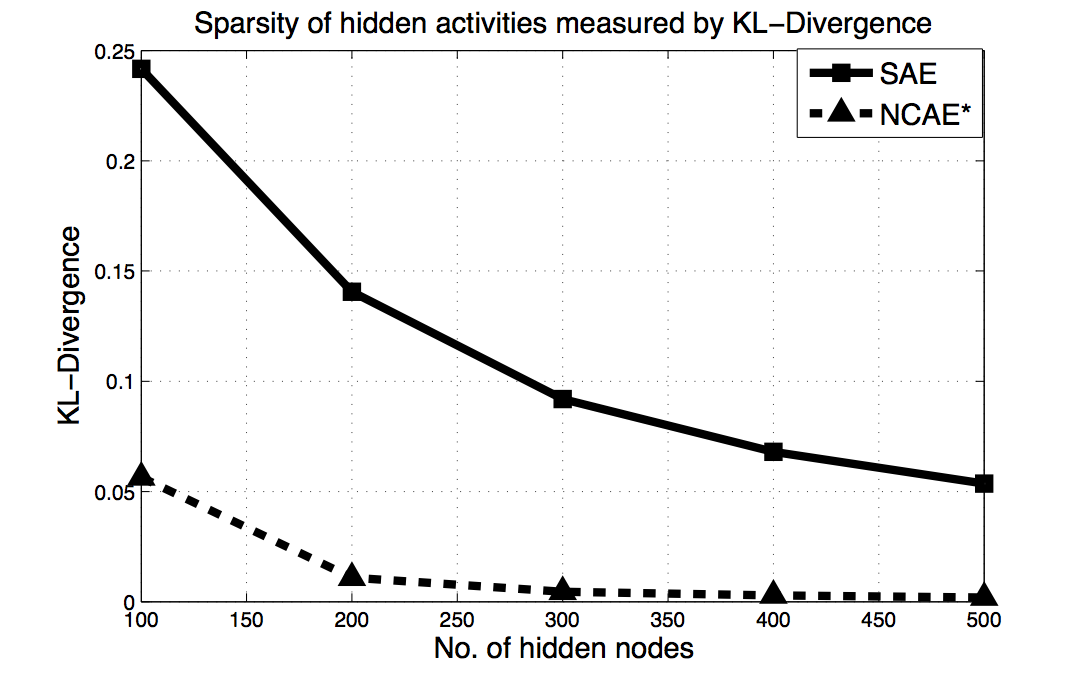}
	\caption{Sparsity of hidden units measured by the KL divergence in Eq. (\ref{eq:cost_kl}) for the MNIST dataset for p$=0.05$.}
	\label{fig:mnist-kl} 
\end{figure}

\begin{figure*}[htb!]
	\begin{minipage}[b]{0.5\linewidth}
		\centering
		\includegraphics[width=9cm]{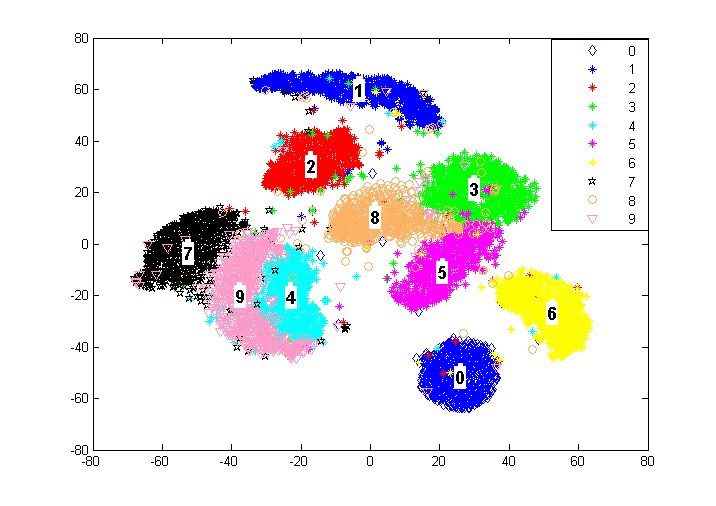}
		{{\footnotesize (a) SAE}}
	\end{minipage}
	\begin{minipage}[b]{0.5\linewidth}
		\centering
		\includegraphics[width=9cm]{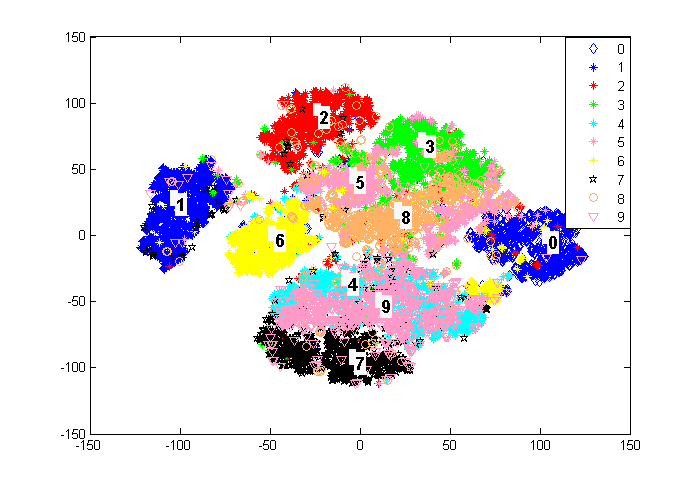}
		{{\footnotesize(b) NNSAE}}
	\end{minipage}
	\begin{minipage}[b]{0.5\linewidth}
		\centering
		\includegraphics[width=9cm]{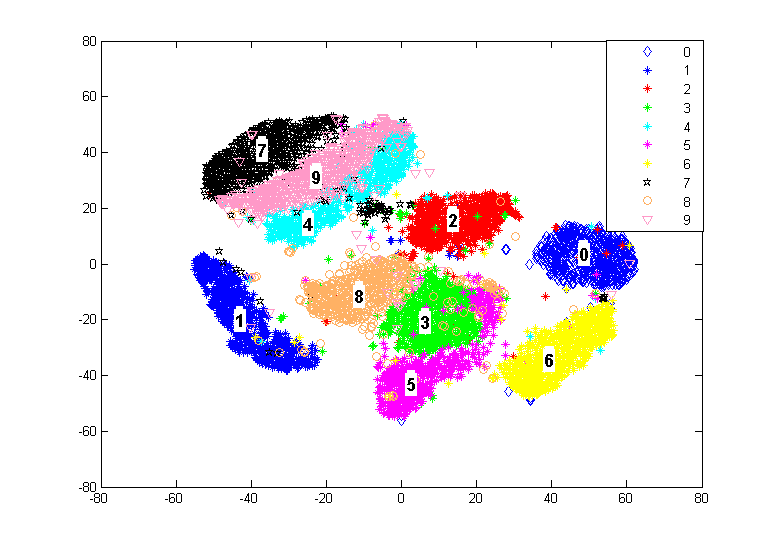}
		{{\footnotesize(c) NCAE*}}
	\end{minipage}
	\begin{minipage}[b]{0.5\linewidth}
		\centering
		\includegraphics[width=9cm]{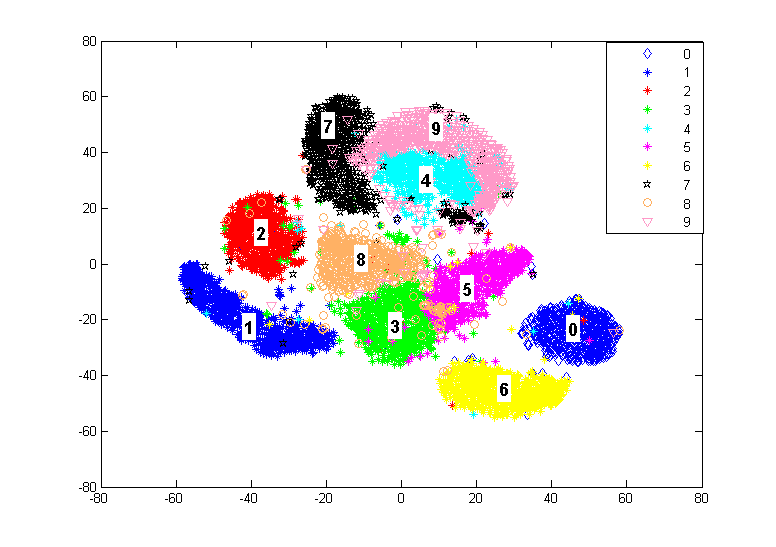}
		{{\footnotesize(d) NMF}}
	\end{minipage}
	
	\caption{Visualization of MNIST handwritten digits. 196 higher representation of digits computed using (a) SAE, (b) NNSAE, (c) NCAE*, and (d) NMF are visualized using t-SNE projection \cite{van2008visualizing}.}
	\label{fig:mnist-visualization} 
\end{figure*}

To better evaluate the hidden activities, Fig. \ref{fig:mnist-kl} depicts the sparsity measured by the KL divergence of Eq. (\ref{eq:cost_kl}) for different numbers of hidden neurons in NCAE and SAE networks. The results 
indicate that the hidden activations in NCAE are more sparse than SAE, since $J_{\text{KL}}\left(p||\hat{\textbf{p}}\right)$ is reduced significantly. This means that the hidden neurons in NCAE are less activated than in SAE when averaged over the full training set. In order to evaluate the ability of the proposed method in discovering the hidden structure of data in the original high-dimensional space, the distributions of MNIST digits in the higher representation level, i.e. hidden activities in SAE, NNSAE and NCAE neural networks, and feature encoding of NMF ($\bf{H}$), are visualized in Fig. \ref{fig:mnist-visualization}(a), \ref{fig:mnist-visualization}(b), \ref{fig:mnist-visualization}(c), and \ref{fig:mnist-visualization}(d) for SAE, NNSAE, NCAE, and NMF, respectively. The figures show the reduced 196-dimensional higher representations of digits in 2D space using t-distributed Stochastic Neighbor Embedding (t-SNE) projection \cite{van2008visualizing}. The comparison between these methods reveals that the distributions of digits for SAE, NCAE, and NMF are more similar to each other than NNSAE. It is clear that manifold of digits in NNSAE have more overlap and more twists than the other methods. On the other hand, the manifolds of digits $7,9,4$ in NCAE are more linear than in SAE and NMF. The comparison between manifolds of other digits in terms of shape and distance indicates that NCAE, SAE, and NMF have similar characteristics.

\begin{figure*}[htb!]
	
	\begin{minipage}[b]{1.0\linewidth}
		\centering
		\includegraphics[width=18cm]{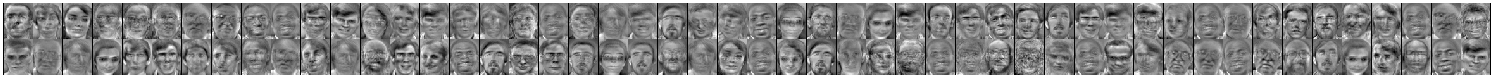}
		\vspace{-2mm}
		{{\footnotesize (a) SAE}}
	\end{minipage}
	\begin{minipage}[b]{1.0\linewidth}
		\centering
		\vspace{2mm}
		\includegraphics[width=18cm]{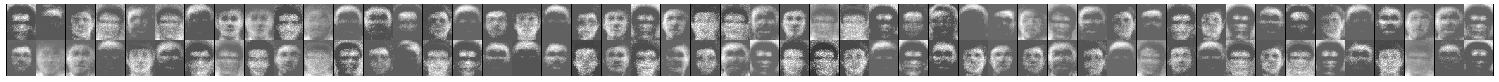}
		\vspace{-2mm}
		{{\footnotesize(b) NNSAE}}
	\end{minipage}
	
	\begin{minipage}[b]{1.0\linewidth}
		\centering
		\vspace{2mm}
		\includegraphics[width=18cm]{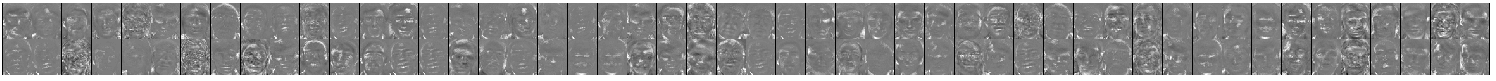}
		\vspace{-2mm}
		{{\footnotesize(c) NCAE*}}
	\end{minipage}
	\begin{minipage}[b]{1.0\linewidth}
		\centering
		\vspace{2mm}
		\includegraphics[width=18cm]{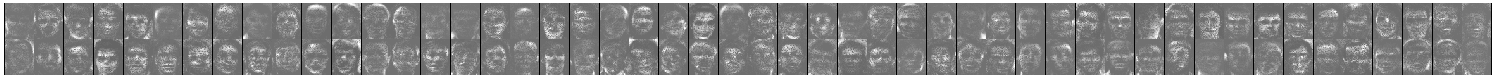}
		\vspace{-2mm}
		{{\footnotesize(d) NMF}}
	\end{minipage}
	\caption{100 Receptive fields learned from the ORL Faces data set using (a) SAE, (b) NNSAE, (c) NCAE*, and (d) NMF. Black pixels indicate negative weights, and white pixels indicate positive weights.}
	\label{fig:ORL-receptivefield}
\end{figure*}

\begin{figure*}[htb!]
	\centering
	\includegraphics[scale=0.7]{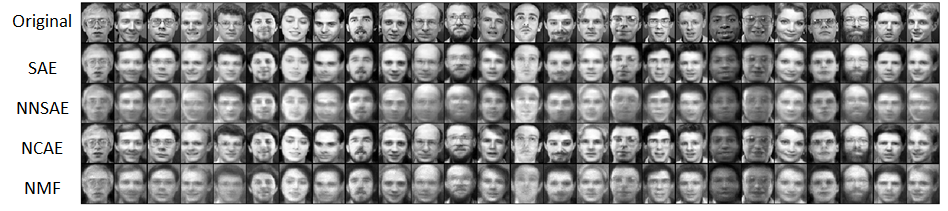}
	\caption{Reconstruction of the ORL Faces test data using 300 receptive fields, using SAE (error=8.6447), NNSAE (error=15.7433), NCAE* (error=5.4944), and NMF (error=7.5653).}\medskip
	\label{fig:ORL-reconstruction}
\end{figure*} 

\begin{figure*}[htb!]
	
	\begin{minipage}[b]{1.0\linewidth}
		\centering
		\includegraphics[width=18cm]{NCAE_ORL.png}
		\vspace{-2mm}
		{{\footnotesize (a) $\alpha=0.003$}}
	\end{minipage}
	\begin{minipage}[b]{1.0\linewidth}
		\centering
		\vspace{2mm}
		\includegraphics[width=18cm]{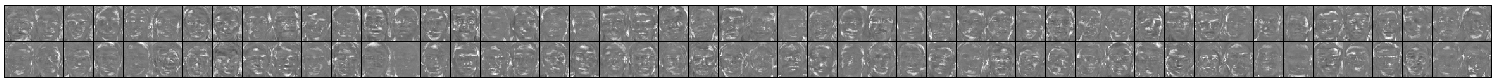}
		\vspace{-2mm}
		{{\footnotesize(b) $\alpha=0.03$}}
	\end{minipage}
	\begin{minipage}[b]{1.0\linewidth}
		\centering
		\vspace{2mm}
		\includegraphics[width=18cm]{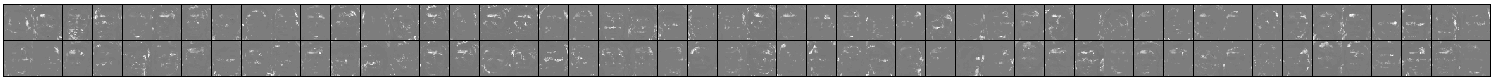}
		\vspace{-2mm}
		{{\footnotesize(c) $\alpha=0.3$}}
	\end{minipage}
	\caption{100 Receptive fields learned from ORL Faces data set using NCAE for varying nonnegativity penalty coefficients $(\alpha)$. Brighter pixels indicate larger weights.}
	\label{fig:ORL-diffparameters}
\end{figure*}

The second experiment is to test the performance of our method on the ORL database of faces (AT\&T at Cambridge) \cite{samaria1994parameterisation}. This database contains $10$ different images of $40$ subjects. For some subjects, the images were taken at different times, with varying lighting, facial expressions, and facial details. The original size of each image is $92\times 112$ pixels, with $256$ gray levels per pixels. To decrease the computational time, we reduce the input layer size of SAE and NCAE by resizing the images to $46\times 56$ pixels. The dataset is divided to $300$ faces for training and $100$ for testing.

The features learned from the ORL data are depicted in the images of receptive fields in Fig. \ref{fig:ORL-receptivefield}(a-d) using the SAE, NNSAE, NCAE, and NMF methods, respectively. The receptive fields of SAE indicate holistic features from different faces, i.e. each feature is a combination of different faces of the database. On the other hand, the receptive fields in NCAE indicate sparse features 
of faces, where several parts of faces can be recognized. Most of the features learned by NCAE contain some parts of the faces, e.g. eye, nose, mouth, etc. together. The nonnegativity constrains negative weights in the NCAE network to become zero, as indicated by fewer darker pixels 
in the receptive fields. The features learned by NNSAE and NMF indicate that most features are holistic, whereas most face parts are visible in the basis images. In NMF and NNSAE, the extracted features are only nonnegative values, but it does not help in creating sparse features, because hard constraints on negative weights force the algorithm to learn complex receptive field of the basis image in NNSAE and NMF, respectively. It can be concluded that
NCAE was able to learn hidden features showing part-based representation of the faces using soft constraints on the negativity weights, whereas this is not achieved by SAE, NNSAE, and NMF. 
To assess the performance of our method in recovering the images, the reconstructed faces of several subjects are shown in Fig. \ref{fig:ORL-reconstruction}. The faces reconstructed by NCAE appear more similar to the original images than those by SAE, NNSAE, and NMF. The reason is that NCAE could extract the hidden features, which show parts of the faces in the encoding layer, and these 
features help the autoencoder network in composing the faces from these features, e.g. eye, nose, mouth, etc., in the decoding layer. However, it is hard to compose the original face from the holistic features created by SAE, NNSAE, and NMF.

\begin{figure*}[htb!]
	
	\begin{minipage}[b]{1.0\linewidth}
		\centering
		\includegraphics[width=18cm]{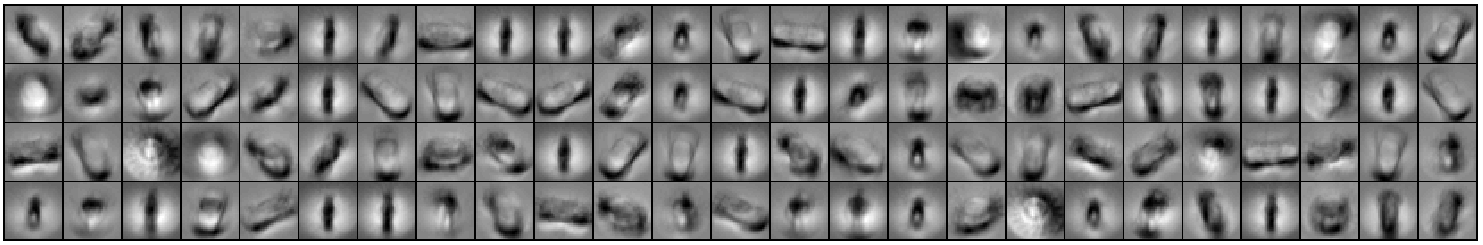}
		\vspace{-2mm}
		{{\footnotesize (a) SAE}}
	\end{minipage}
	\begin{minipage}[b]{1.0\linewidth}
		\centering
		\vspace{2mm}
		\includegraphics[width=18cm]{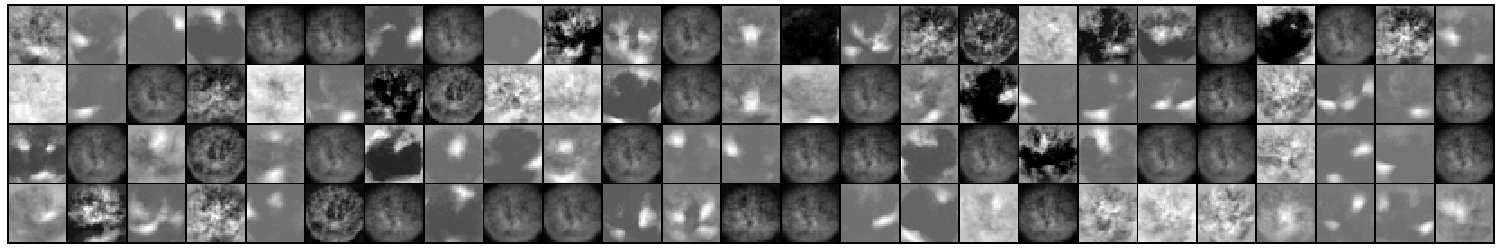}
		\vspace{-2mm}
		{{\footnotesize(b) NNSAE}}
	\end{minipage}
	\begin{minipage}[b]{1.0\linewidth}
		\centering
		\vspace{2mm}
		\includegraphics[width=18cm]{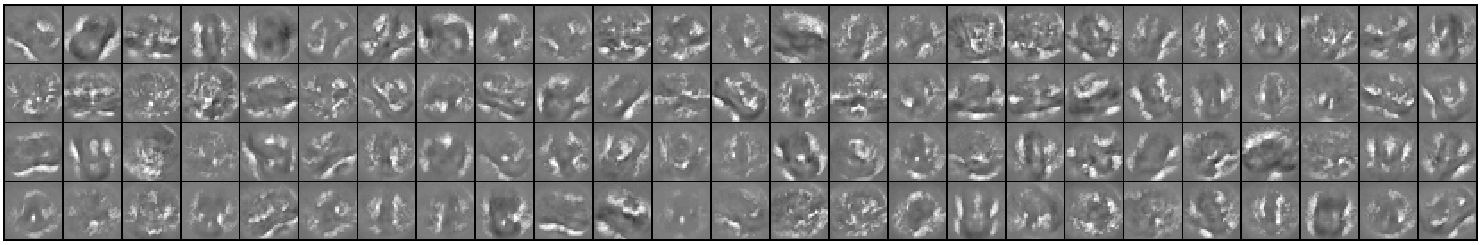}
		\vspace{-2mm}
		{{\footnotesize(c) NCAE*}}
	\end{minipage}
	\begin{minipage}[b]{1.0\linewidth}
		\centering
		\vspace{2mm}
		\includegraphics[width=18cm]{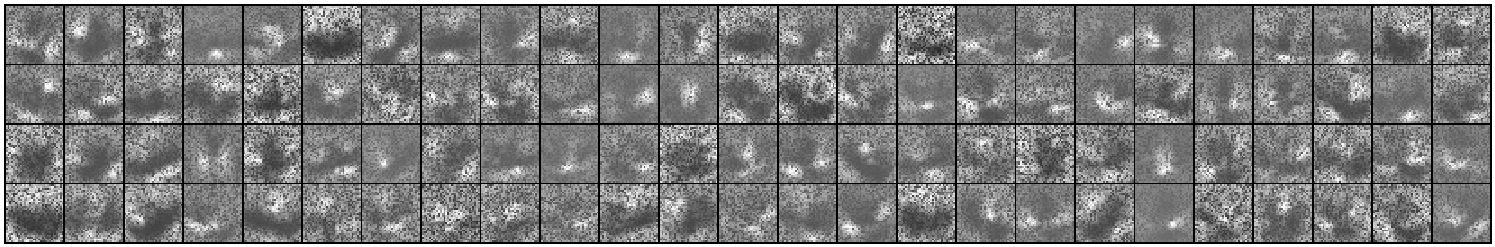}
		\vspace{-2mm}
		{{\footnotesize(d) NMF}}
	\end{minipage}
	\caption{100 Receptive fields learned from small NORB data set using (a) SAE, (b) NNSAE, (c) NCAE*, and (d) NMF. Black pixels indicate negative, and white pixels indicate positive weights.}
	\label{fig:NORB-receptivefields}
\end{figure*}

To investigate the effect of the nonnegativity constraint penalty coefficient ($\alpha$) in NCAE for learning part-based representation, 
we test different values of $\alpha$ to train NCAE. The hidden features are depicted in Fig. \ref{fig:ORL-diffparameters}. For this experiment, we increase $\alpha$ logarithmically for 
$3$ values in the range $[0.003,\ldots,0.3]$. The results indicate that by increasing $\alpha$, the resulting features are more sparse, and decompose faces into smaller parts. It is clear that the receptive fields in Fig. \ref{fig:ORL-diffparameters}(c) are more sparse, and only show few parts of the faces. This test demonstrates that NCAE is able to extract different types of eyes, noses, mouths, etc. from the face database.

In the third experiment, we use the NORB normalized-uniform dataset \cite{lecun2004learning}, which
contains $24,300$ training examples and $24,300$ test examples. This database contains images 
of $50$ toys from $5$ generic categories: four-legged animals, human figures, airplanes, trucks, and cars. The training and testing sets are composed of $5$ instances of each category. Each image consists of two channels, each of size $96\times 96$ pixels. We take the inner $64\times 64$ pixels of each channel and resize it using bicubic interpolation to $32\times 32$ pixels that form a vector with $2048$ entries as the input. To evaluate the performance of the method, we train an autoencoder using $100$ hidden neurons for SAE, NNSAE, and NCAE, and also NMF with $100$ basis vectors. The learned features are shown as receptive fields in Fig. \ref{fig:NORB-receptivefields}. The results indicate that the receptive fields learned by NCAE are more sparse than SAE and NNSAE, since they mainly capture the edges of the toys. On the other hand, the receptive fields from SAE and NNSAE represent more holistic features. The basis images learned by NMF also show edge-like features, however, they are more holistic than the NCAE features.

\begin{figure*}[htb!]
	\centering
	\includegraphics[scale=0.5]{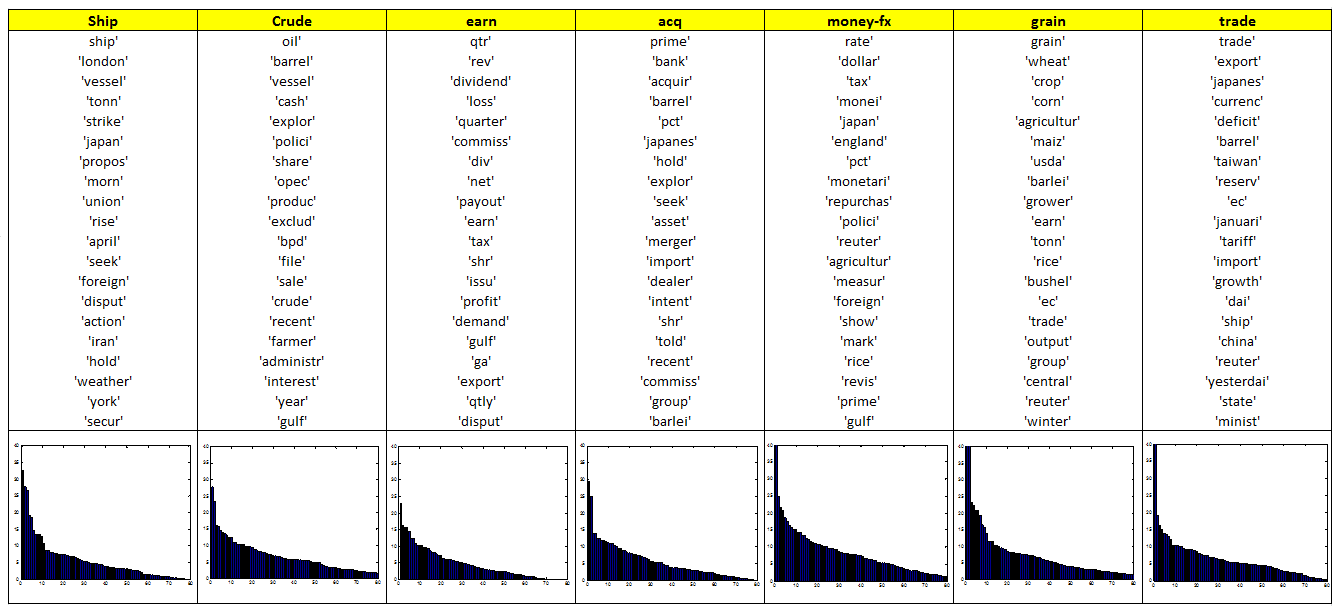}
	\caption{An example of 7 most distinguished categories, i.e., ship, crude, earn, acq, money-fx, grain and trade associated with top 20 words (ranked by their weights) discovered
		from the Reuters-21578 document corpus. The charts at the bottom row illustrate the weight impact of words on the category (x axis refers to words and y axes refers to theirs weights). These weights are sorted in descending order.}\medskip
	\label{fig:reuters-topics}
\end{figure*} 

\begin{figure*}[htb!]
	\begin{minipage}[b]{0.5\linewidth}
		\centering
		\includegraphics[width=9cm]{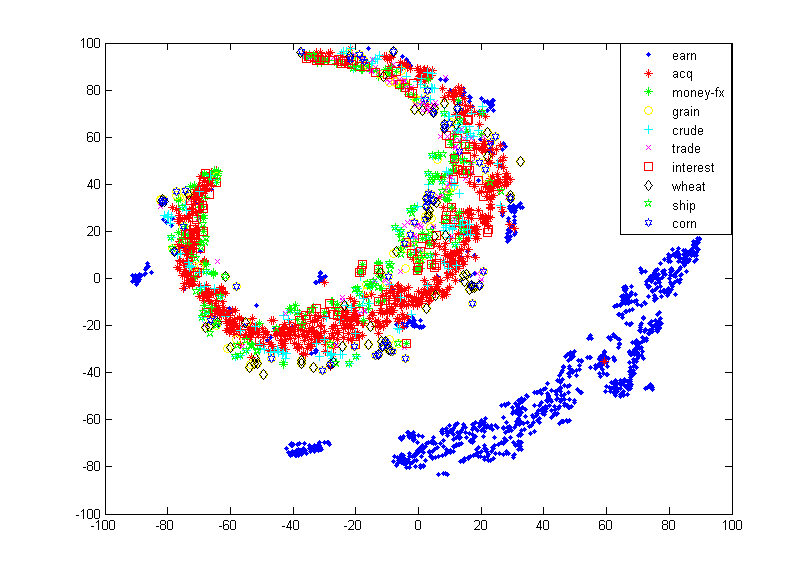}
		{{\footnotesize (a) SAE}}
	\end{minipage}
	\begin{minipage}[b]{0.5\linewidth}
		\centering
		\includegraphics[width=9cm]{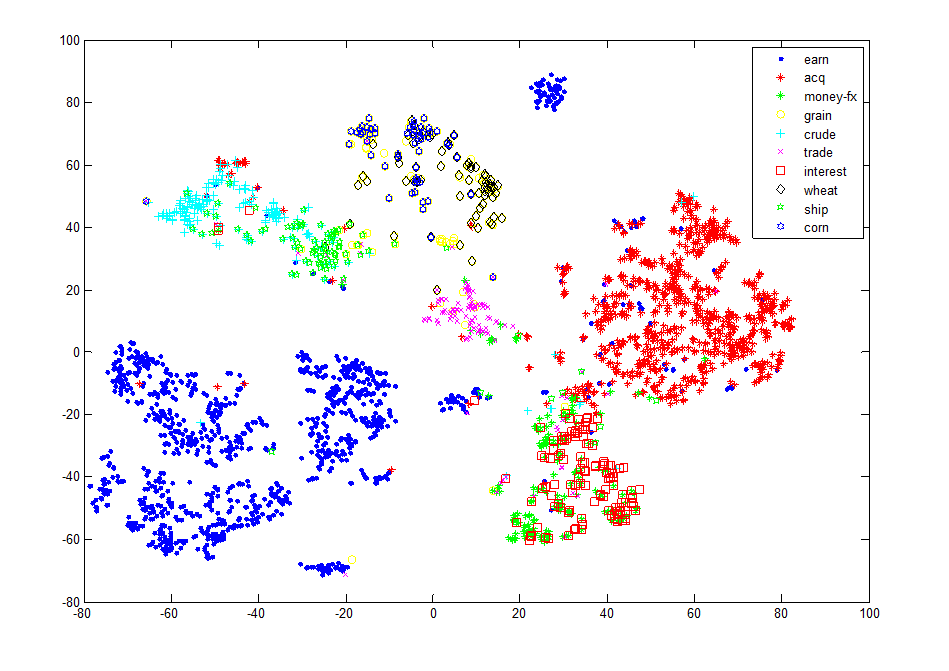}
		{{\footnotesize(b) NCAE*}}
	\end{minipage}
	
	\caption{Visualization of the Reuters documents data based on the 15-dimensional higher representation of documents computed using (a) SAE and (b) NCAE*. Visualization used t-SNE projection \cite{van2008visualizing}.}
	\label{fig:reuters-visualization} 
\end{figure*}

\subsubsection{Semantic feature discovery from text data}
In this experiment, the NCAE method is evaluated on extracting semantic features from text data. The documents are first converted to a TF-IDF vector space model~\cite{sparck1972statistical}. Part-based representation in text documents is more complicated than in images, since the meaning of document can not be easily inferred by adding the words it contains. However, the topic of a document can be inferred from a group of words delivering the most information. Therefore, to detect the topic of a document, the autoencoder network should be able to extract these groups of words in its encoding layer to generate a meaningful semantic embedding. To evaluate our proposed method, we used the Reuters-21578 text categorization collection. It is composed
of documents that appeared in the Reuters newswire
in 1987. We used the ModApte split limited to 10 most
frequent categories. We have used a processed (stemming,
stop-word removal) version in bag-of-words format obtained
from http://people.kyb.tuebingen.mpg.de/pgehler/rap/.

The dataset contains $11,413$ documents with $12,317$ words/dimensions. Two techniques were used to reduce the dimensionality of each document to contain the most informative and less correlated words. First, words were sorted based on their frequency of occurrence in the dataset. Then the words with frequency below 4 and above $70$ were removed. After that, the information gain with the class attribute \cite{pang2006introduction} was used to select the most informative words which do not occur in every topic. The remaining words in the dataset were sorted using this method, and the less important words were removed based on the desired dimension of documents. In this experiment, we reduced the dimensionality of documents to the size $[200, 300, 400]$.

To examine the features extracted in the encoding layer of NCAE, $20$ words connecting via the highest weights to each hidden neuron were examined. The connecting weight from each word to a hidden neuron is equal to the magnitude of the association of the word to the latent feature extracted by the hidden node. Using this interpretation, a hidden node with the largest connecting weight of words related to a specific topic can be assigned as a class detector for that topic. We train an NCAE network with $200$ input neurons and $15$ hidden neurons. Fig. \ref{fig:reuters-topics} depicts the selected seven nodes showing the seven distinguishable topics of the dataset. The top row shows the list of words with the topic inferred from the corresponding list. The bottom row depicts the distribution of connecting weights in decreasing order. It can be concluded that the semantically related words  of a topic are grouped together in each hidden node. To further evaluate the ability of the NCAE network to disentangle the semantic features (topic detector) from the dataset, the distribution of documents in the hidden layer is compared to the SAE method, as depicted in Fig. \ref{fig:reuters-visualization}, where topic information is used for visual labeling. It is clear that NCAE is able to group the related documents together, whereas the topics which are meaningfully related are closer in the semantic space.


\begin{figure*}[htb!]
	\begin{minipage}[b]{1.0\linewidth}
		\centering
		\includegraphics[width=18cm]{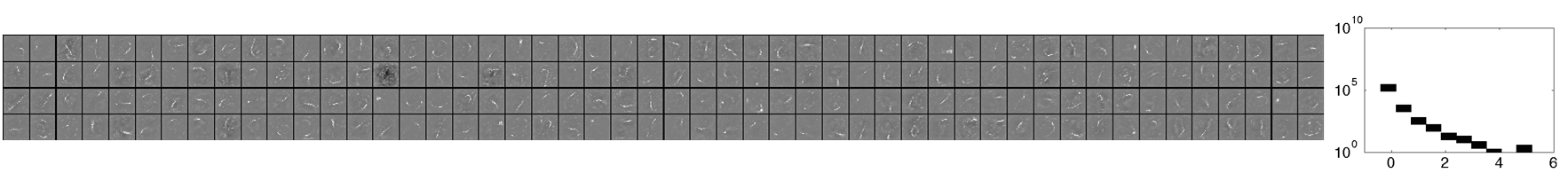}
		\vspace{-2mm}
		{{\footnotesize (a)}}
	\end{minipage}
	\begin{minipage}[b]{1.0\linewidth}
		\centering
		\vspace{2mm}
		\includegraphics[width=18cm]{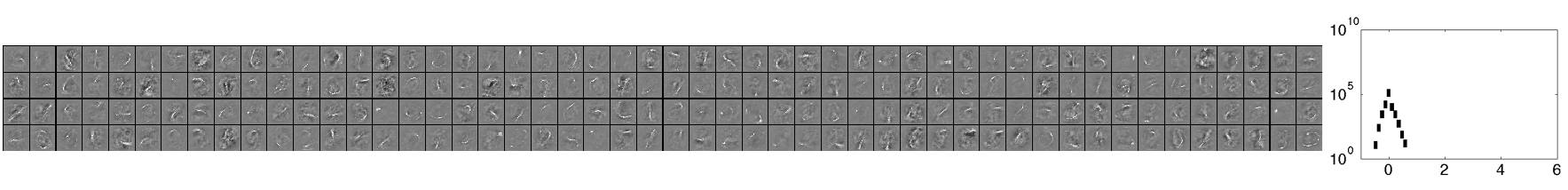}
		\vspace{-2mm}
		{{\footnotesize(b)}}
	\end{minipage}
	\caption{200 Receptive fields of the first layer of the deep network after fine-tuning using (a) all weights constrained, and (b) only Softmax weights constrained. According to histogram, $5.76\%$ of weights become negative. Black pixels indicate negative weights, and white pixels indicate positive weights.}
		\label{fig:mnist-RF-after-constraint}
\end{figure*}

\begin{figure*}[htb!]
	\begin{minipage}[b]{1.0\linewidth}
		\centering
		\includegraphics[width=18cm]{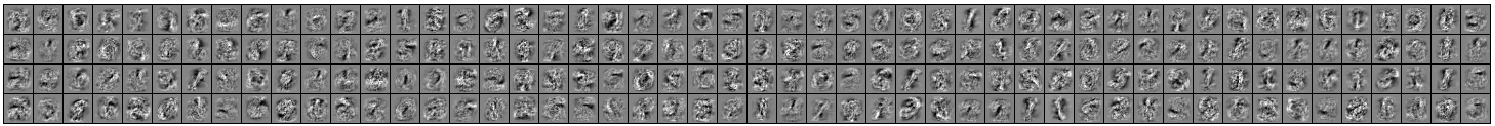}
		\vspace{-2mm}
		{{\footnotesize (a) SAE}}
	\end{minipage}
	\begin{minipage}[b]{1.0\linewidth}
		\centering
		\vspace{2mm}
		\includegraphics[width=18cm]{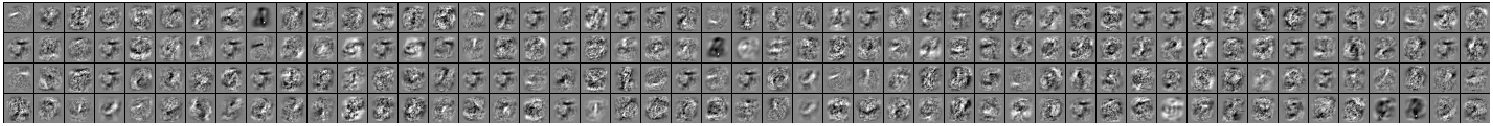}
		\vspace{-2mm}
		{{\footnotesize(b) NNSAE}}
	\end{minipage}
	\begin{minipage}[b]{1.0\linewidth}
		\centering
		\vspace{2mm}
		\includegraphics[width=18cm]{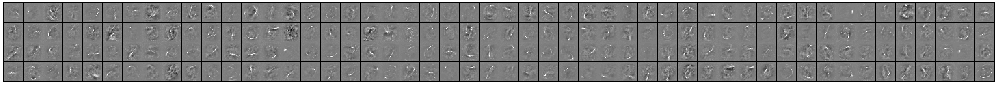}
		\vspace{-2mm}
		{{\footnotesize(c) NCAE*}}
	\end{minipage}
	\begin{minipage}[b]{1.0\linewidth}
		\centering
		\vspace{2mm}
		\includegraphics[width=18cm]{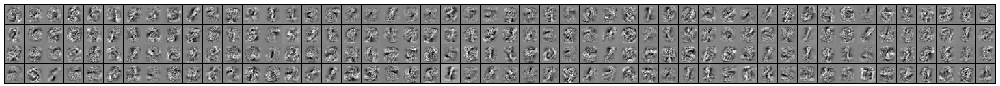}
		\vspace{-2mm}
		{{\footnotesize(d) DAE}}
	\end{minipage}
	\begin{minipage}[b]{1.0\linewidth}
		\centering
		\vspace{2mm}
		\includegraphics[width=18cm]{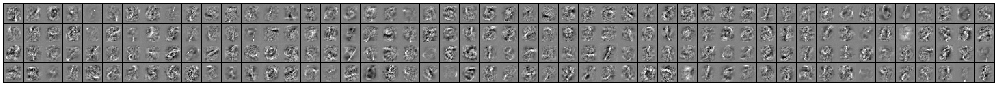}
		\vspace{-2mm}
		{{\footnotesize(e) NC-DAE}}
	\end{minipage}
	\begin{minipage}[b]{1.0\linewidth}
		\centering
		\vspace{2mm}
		\includegraphics[width=18cm]{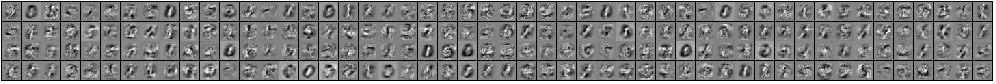}
		\vspace{-2mm}
		{{\footnotesize(f) DpAE}}
	\end{minipage}
	\caption{200 Receptive fields of the first layer of the deep network after fine-tuning using (a) SAE, (b) NNSAE, (c) NCAE*, (d) DAE, (e) NC-DAE, and (f) DpAE on the MNIST data. Black pixels indicate negative, and white pixels indicate positive weights.}
	\label{fig:mnist-RF-after}
\end{figure*}

\subsection{Supervised Learning}

The next step is to investigate whether the ability of a deep network in to decompose data into parts, with improved ability to disentangle the hidden factors in its layers, can improve prediction performance. In this section, we pre-train a deep network by stacking several NCAE networks, trained in the previous section. Then a softmax classifier is trained using the hidden activities of the last autoencoder using Eq. (\ref{eq:cost-NCsoftmax}). Finally, the deep network is fine-tuned using Eq. (\ref{eq:cost-FineTuning}) to improve the classification accuracy. The results are compared to deep neural networks trained using SAE, NNSAE, Denoising Autoencoder (DAE) \cite{vincent2008extracting}, and Dropout Autoencoder (DpAE) \cite{hinton2012improving} on the MNIST, NORB, and the Reuters-21578 text corpus datasets. The classification results are averaged over $10$ experiments to mitigate the effect of initialization.

Tables \ref{table:MNIST-classification}-\ref{table:Reuters200-Results} report the classification accuracy of the deep network, pre-trained with different autoencoders. The results indicate that a deep network with NCAE yields a significantly better accuracy than other networks \textit{before} fine-tuning for all three datasets, and \textit{after} fine-tuning for two of the three data sets. For the NORB data set, although the NCAE network was significantly superior \textit{before} fine-tuning, the classification results
indicate no significant difference between NCAE, DAE, and DpAE networks, \textit{after} fine-tuning.
The convergence speed of the different networks were also compared based on the number of iterations during fine-tuning. These are listed alongside the error rates in Tables~\ref{table:MNIST-classification}-\ref{table:Reuters200-Results}. It can be seen that NCAE network converges faster than other methods, since it yields better accuracy \textit{before} fine-tuning. Note that all networks were trained for the same number of iterations $(400)$ \textit{before} fine-tuning. Therefore NCAE's superior performance is not at the cost of more iterations.

\begin{table}[htb!]
	\caption{Classification performance of supervised learning methods on
		MNIST dataset.}
	\scalebox{0.65}{
		\begin{tabular}{|l|c c|c c c|}
			\hline
			\multicolumn{1}{|c|}{} & \multicolumn{2}{c|}{\textit{Before} fine-tuning} & \multicolumn{3}{|c|}{\textit{After} fine-tuning} \\
			\hline
			\multicolumn{1}{|c|}{Model (784-200-20-10)} & Mean$\pm$ SD & \textit{p}-value & Mean$\pm$ SD & \textit{p}-value & \# Iterations \\
			\hline
			Deep NCAE* & $\textbf{84.83}\pm \textbf{0.094}$ & &$\textbf{97.91}\pm \textbf{0.1264}$ & & $\textbf{97}$\\
			Deep SAE& $52.81\pm 0.1277$ &$\textless 0.0001$ & $97.29\pm 0.091$ & $\textless 0.0001$ & $400$\\
			Deep NNSAE & $69.72\pm 0.1007$ & $\textless 0.0001$& $97.18\pm 0.0648$ &$\textless 0.0001$ & $400$\\
			Deep DAE ($50\%$ input dropout) & $11.26\pm 0.14$ & $\textless 0.0001$& $97.11\pm 0.0808$ & $\textless 0.0001$ & $400$\\
			Deep NC-DAE ($50\%$ input dropout) & $84.37\pm 0.1318$ &$\textless 0.0001$ & $97.42\pm 0.0757$& $\textless 0.0001$ & $106$\\
			Deep DpAE ($50\%$ hidden dropout) & $16.77\pm 0.0784$ & $\textless 0.0001$ & $96.73\pm 0.1066$ & $\textless 0.0001$ & $400$ \\
			\hline
		\end{tabular}
		}
		\label{table:MNIST-classification}
	\end{table}
	
	\begin{table}[htb!]
		\caption{Classification performance of supervised learning methods on
			NORB dataset.}
		\scalebox{0.65}{
			\begin{tabular}{|l|c c|c c c|}
				\hline
				\multicolumn{1}{|c|}{} & \multicolumn{2}{c|}{\textit{Before} fine-tuning} & \multicolumn{3}{|c|}{\textit{After} fine-tuning} \\
				\hline
				\multicolumn{1}{|c|}{Model (2048-200-20-5)} & Mean$\pm$ SD & \textit{p}-value & Mean$\pm$ SD & \textit{p}-value & \# Iterations\\
				\hline
				Deep NCAE* & $\textbf{75.54}\pm \textbf{0.1152}$ & & $87.76\pm 0.3613$ & & $\textbf{242}$\\
				Deep SAE& $20.00\pm 0.1768$ & $\textless 0.0001$& $87.26\pm 0.3109$ & $0.0039 $& $400$\\
				Deep NNSAE & $19.93\pm 0.2230$ & $\textless 0.0001$& $79.00\pm 0.0962$ & $\textless 0.0001$& $400$\\
				Deep DAE ($50\%$ input dropout) & $44.03\pm 0.1553$ & $\textless 0.0001$& $\textbf{88.11}\pm \textbf{0.3861}$ & $0.0508$ & $400$\\
				Deep DpAE ($50\%$ hidden dropout) & $49.49\pm 0.1437$ & $\textless 0.0001$ & $87.75\pm 0.2767$ & $0.9454$ & $400$\\
				\hline
			\end{tabular}
			}
			\label{table:norb2048-Results}
		\end{table}
		
		\begin{table}[htb!]
			\caption{Classification performance of supervised learning methods on
				Reuters-21578 Dataset.}
			\scalebox{0.65}{
				\begin{tabular}{|l|c c|c c c|}
					\hline
					\multicolumn{1}{|c|}{} & \multicolumn{2}{c|}{\textit{Before} fine-tuning} & \multicolumn{3}{|c|}{\textit{After} fine-tuning} \\
					\hline
					\multicolumn{1}{|c|}{Model (200-15-10)} & Mean$\pm$ SD & \textit{p}-value & Mean$\pm$ SD & \textit{p}-value & \# Iterations\\
					\hline
					Shallow NCAE* & $\textbf{57.18}\pm \textbf{0.3639}$& & $\textbf{81.15}\pm \textbf{0.1637}$ & & $400$\\
					Shallow SAE & $39.00\pm 0.2255$& $\textless 0.0001$& $78.60\pm 0.2143$&$\textless 0.0001$& $400$\\
					Shallow DAE ($50\%$ input dropout) & $39.00\pm 0.3617$& $\textless 0.0001$& $76.35\pm 0.1918$&$\textless 0.0001$& $400$\\
					Shallow DpAE ($20\%$ hidden dropout) & $39.00\pm 0.4639$& $\textless 0.0001$& $78.04\pm 0.1709$& $\textless 0.0001$& $400$\\
					Shallow DpAE ($50\%$ hidden dropout) & $39.00\pm 0.3681$& $\textless 0.0001$& $72.12\pm 0.2901$& $\textless 0.0001$& $400$\\
					\hline
				\end{tabular}
				}
				\label{table:Reuters200-Results}
			\end{table}

			To relate the improved accuracy to part-based decomposition of data, the first hidden layer of each deep network is depicted in Fig. \ref{fig:mnist-RF-after}. It demonstrates that the deep network based on NCAE could decompose data into clearly distinct parts in the first layer, whereas there are more holistic features in other networks. This property leads to better discrimination between classes at the next layers, thus resulting in better classification. The nonnegativity constraint in the DAE network (NC-DAE) has also been tested on the MNIST dataset. The results indicate that the performance of NC-DAE improves over DAE \textit{before} and \textit{after} fine-tuning, since the hidden layers of NC-DAE are able to decompose data into parts, and also it converges faster \textit{after} fine-tuning. Table \ref{table:norb2048-Results} reports the classification results on the small NORB dataset, and demonstrates that the deep network with NCAE outperforms the other networks \textit{before} fine-tuning. However, its performance is not significantly different from the deep networks based on DAE and DpAE, based on the \textit{p}-values. Table \ref{table:Reuters200-Results} also reports the classification accuracy computed with several one-hidden-layer networks. It also indicates that the deep network built with NCAE outperforms other deep networks \textit{before} and \textit{after} fine-tuning by a large margin, due to their ability to extract semantic features from documents on this data.

			\section{Conclusion}
			In this paper, a new learning algorithm was proposed for training a deep autoencoder-based network with nonnegative weights constraints, first in the unsupervised training
			of the autoencoder (NCAE), and then in the supervised fine-tuning stage. Nonnegativity has been motivated by the idea in NMF that it promotes additive features and captures part-based representation of data. The performance of the proposed method, in terms of decomposing data into parts and extracting meaningful features, was compared to the Sparse Autoencoder (SAE), Nonnegative Sparse Autoencoder (NNSAE) and NMF. The prediction performance of the deep network has also been compared to the SAE, NNSAE, Denoising Autoencoder (DAE), and Dropout Autoencoder (DpAE). We evaluated the performance on the MNIST data set of handwritten digits, the ORL data set of faces, small NORB data set of objects, and Reuters-21578 text corpus. The results were evaluated in terms of reconstruction error, part-based representation of features, and sparseness of hidden encoding in the unsupervised learning stage. The results indicate that the nonnegativity constraints, in the NCAE method, force the autoencoder to learn features that capture a part-based representation of data, while achieving lower reconstruction error and better sparsity in the hidden encoding as compared with SAE and NMF. The numerical classification results also reveal that a deep network trained by restricting the number of nonnegative weights in the autoencoding and softmax classification layer achieves better performance. This is due to decomposing data into parts, hierarchically in its hidden layers, which 
			help the classification layer to discriminate between classes.


			\section*{Acknowledgment}
			This work is partially supported by KSEF-3113-RDE-017.

\bibliographystyle{IEEEtran}

\bibliography{references}

\vspace{-12mm}
\begin{IEEEbiography}[{\includegraphics[width=1in,height=1.25in,clip,keepaspectratio]{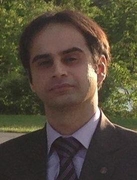}}]
{Ehsan Hosseini-Asl} (M'12) received the M.Sc. degree in electrical engineering (Automation and Instrumentation Systems) from the Petroleum University of Technology, Tehran, Iran. Since 2014 he is a Ph.D. candidate at the University of Louisville, Kentucky, USA.

He authored or co-authored six conference papers published in the proceedings, two journal papers, and several papers in submission. His current research interests include unsupervised feature learning and deep learning techniques and their applications representation learning of 2D and 3D images, biomedical image segmentation, tissue classification and disease detection. 

In 2012, he was the recipient of the University Scholarship from the University of Louisville where he now serves as Research and Teaching Assistant. He was also the recipient of Graduate Student Research grant from IEEE Computational Intelligence Society (CIS) in 2015 for his deep learning research. 
\end{IEEEbiography} 

\vspace{-12mm}
\begin{IEEEbiography}[{\includegraphics[width=1in,height=1.25in,clip,keepaspectratio]{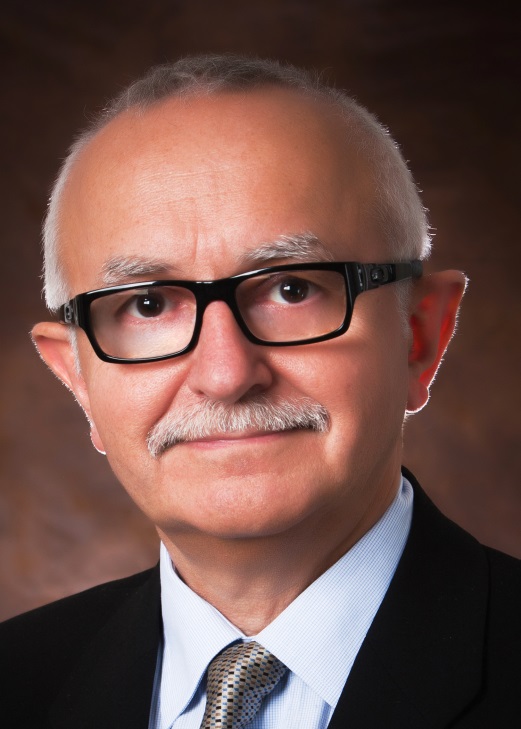}}]
{Jacek M. Zurada} 
(M'82-SM'83-F'96-LF’14) Ph.D., has received his degrees from Gdansk Institute of Technology, Poland. He now serves as a Professor of Electrical and Computer Engineering at the University of Louisville, KY. He authored or co-authored several books and over 380 papers in computational intelligence, neural networks, machine learning, logic rule extraction, and bioinformatics, and delivered over 100 presentations throughout the world. His work has been cited over 8500 times. 

In 2014 he served as IEEE V-President, Technical Activities (TAB Chair). In 2015 he chairs the IEEE TAB Strategic Planning Committee.  He chaired the IEEE TAB Periodicals Committee, and TAB Periodicals Review and Advisory Committee, respectively, and was the Editor-in-Chief of the IEEE Transactions on Neural Networks (1997-03), Associate Editor of the IEEE Transactions on Circuits and Systems, Neural Networks and of The Proceedings of the IEEE. In 2004-05, he was the President of the IEEE Computational Intelligence Society.  

Dr. Zurada is an Associate Editor of Neurocomputing, and of several other journals. He holds the title of a Professor in Poland and is a member of the Polish Academy of Sciences. He has been awarded numerous distinctions, including the 2013 Joe Desch Innovation Award and 2015 Distinguished Service Award, and four honorary professorships.  He has been a Board Member of such professional associations as the IEEE, IEEE CIS and IJCNN.    
\end{IEEEbiography}

\vspace{-12mm}
\begin{IEEEbiography}[{\includegraphics[width=1in,height=1.25in,clip,keepaspectratio]{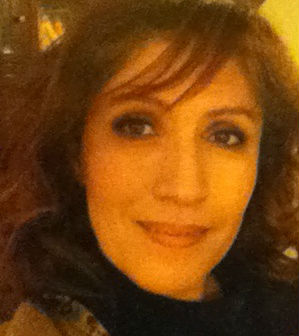}}]
{Olfa Nasraoui} (SM'15) 
is a Professor in Computer Science and Engineering at the University of Louisville, where she is also  endowed Chair of e-commerce and the founding director of the Knowledge Discovery \& Web Mining Lab. She received her Ph.D in Computer Engineering and Computer Science from the University of Missouri-Columbia in 1999. Her research activities include data mining, in particular, Web mining, mining evolving data streams, and Web personalization. She has served on the organizing and program committees of several conferences and workshops, including organizing the premier series of workshops on Web Mining, WebKDD 2004-2008. She has also served as Program Committee Vice-Chair, Track Chair, or Senior Program Committee member for several data mining conferences including KDD, ICDM, WWW, AAAI, SDM, RecSys, and CIKM.

She is the recipient of the National Science Foundation CAREER Award and her research has been funded mainly by NSF and NASA. She has more than 160 publications in journals, book chapters and refereed conferences, as well as 10 edited volumes. She is a member of IEEE, IEEE CIS, IEEE WIE, ACM, and ACM SIG-KDD.
\end{IEEEbiography}

\end{document}